\documentclass[sigconf]{acmart}

\copyrightyear{2026}
\acmYear{2026}
\setcopyright{cc}
\setcctype{by-nc-nd}
\acmConference[CIKM '26]{Proceedings of the 35th ACM International Conference on Information and Knowledge Management}{November 7--11, 2026}{Rome, Italy.}
\acmBooktitle{Proceedings of the 35th ACM International Conference on Information and Knowledge Management (CIKM '26), November 7--11, 2026, Rome, Italy}
\acmISBN{979-8-4007-2539-5/2026/11}
\acmDOI{10.1145/3799682.3839256}
\usepackage{booktabs}
\usepackage{array}
\usepackage{algorithm}
\usepackage{algpseudocode}
\usepackage{tablefootnote}
\usepackage{balance}

\begin{document}

\title{Optimize the Harness, Not the Model: Fitting a Minimal Agent to a
Self-Hostable LLM for Investigative Analysis}

\author{Shai Nahum Gefen}
\affiliation{%
  \institution{Dream Security}
  \city{Tel Aviv}
  \country{Israel}
}
\email{shai@dreamgroup.com}

\author{Vered Chassida}
\affiliation{%
  \institution{Dream Security}
  \city{Tel Aviv}
  \country{Israel}
}
\email{veredch@dreamgroup.com}

\author{Dmitry Basin}
\affiliation{%
  \institution{Dream Security}
  \city{Tel Aviv}
  \country{Israel}
}
\email{dmitryba@dreamgroup.com}

\author{Guy Feigenblat}
\affiliation{%
  \institution{Dream Security}
  \city{Tel Aviv}
  \country{Israel}
}
\email{guyfe@dreamgroup.com}

\renewcommand{\shortauthors}{Shai Nahum Gefen, Vered Chassida, Dmitry Basin, and Guy Feigenblat}

\begin{abstract}
Government intelligence teams need LLM agents that can analyze large,
messy evidence collections, but security and cost constraints often rule
out hosted frontier models and broad, hard-to-audit tool surfaces. We
report \emph{CLEO}: \textbf{C}onstrained \textbf{L}oop for \textbf{E}nvironment \textbf{O}ptimization, a
method for fitting a minimal agent harness to a fixed, self-hostable LLM.
Given an expert-validated dataset, CLEO admits only the tools,
guardrails, memory policies, and loop components that improve accuracy
while remaining general and auditable.
On our held-out dataset, the CLEO harness reaches an $83.8$\% pass rate, within $1.2$ points of the strongest out-of-the-box Claude Code SDK baseline, while being two orders of magnitude smaller.
\end{abstract}




\maketitle

\section{Introduction}
Analysts in government organizations investigate complex cases spanning
thousands of heterogeneous documents---emails, reports, call records,
spreadsheets, chat logs, and multilingual material with inconsistent or unknown
schemas---revisiting them repeatedly to aggregate evidence, reconcile
contradictions, trace entities, and build coherent explanations.

LLM agents can make such investigations tractable, but only through a
\emph{harness}: the tools, memory, execution policies, and control loop
that let the model search corpora, inspect files, run analyses, and
refine hypotheses. In high-security settings both the served model and the harness are
constrained: data sensitivity and cost push organizations toward smaller
self-hostable models, while security review forces tight control over exposed
capabilities---every new tool and every line of harness code must be inspected,
justified, and often certified before deployment.

Two lines of work bear on this. \emph{Harness design} spans
reason-and-act control \cite{yao2023react}, filesystem agent--computer
interfaces \cite{yang2024sweagent}, lean and generic-purpose harness implementations such as
PI~\cite{pi}.
\emph{Harness optimization}
automates this design, from agent-design search \cite{hu2025adas} to
autoresearch-like loops \cite{karpathy2026autoresearch} and
observability-driven harness evolution \cite{lin2026ahe}. Evidence access is a further design choice: where the dominant pattern
pre-indexes the corpus into embeddings \cite{lewis2020rag},
\emph{indexless} access through raw-file search and scripts can match or
beat strong retrievers on evolving, heterogeneous corpora \cite{li2026dci}.

We report \emph{CLEO}, a method that enabled us to launch a \emph{deployed}
investigative agent around a fixed, self-hostable model and a deliberately
minimal harness. We use CLEO for both the optimization method and the resulting production harness. CLEO's first contribution is an explicit
\textbf{justification gate}: proposed harness changes must improve
accuracy and also be generic, auditable, and worth the added capability
surface. This realizes the multi-objective harness that was mentioned but not studied in \cite{hu2025adas}. Its second contribution is \textbf{bottom-up harness
construction}: the harness starts nearly empty and grows only when
evaluation shows that a new tool, guardrail, memory policy, or loop
component is necessary. Minimality is therefore optimized by construction,
not patched in later. The resulting agent holds up on held-out,
out-of-distribution cases (Table~\ref{tab:harness}), is small enough for
security review, and is currently being deployed in client environments.

\section{Method}
\textbf{Datasets.} Intelligence analysts work across distinct \emph{use cases},
each a self-contained investigation with its own documents and questions; the
product must serve many, so we build and evaluate on three. Within a use case,
files are grouped into \emph{corpora}---ingestion, classification,
investigation, insights, reporting, and multilingual content---that overlap in
their documents but differ in which files are relevant to a given query. A
representative use case holds about 8{,}000 files across up to seven formats,
with corpora from roughly 250\,MB to 6\,GB, and includes misleading and
contradictory evidence to stress robustness. Across the three use cases we
author 300+ expert questions at four difficulty levels; 240 are held out for evaluation and the rest
form the optimization pool.

\subsection{Model, Data, and a Minimal Harness}

\textbf{The model.}
We work with self-hostable (on-premises), open-weights LLMs and treat the model as
\emph{fixed}: the method optimizes everything around it.
We report results with GLM-5.1~\cite{glm5}.

\textbf{Optimization-loop dataset.}
The optimization loop runs on a single use case of four corpora, approximately
3{,}000 files in total. Each iteration evaluates the harness on 44 Q\&A pairs:
a fixed core of 32---eight per corpus, drawn from all difficulty levels---plus
12 sampled fresh from the development pool every iteration. We keep the evaluation set small so each optimization iteration finishes within minutes, staying inside our time and compute budget. Periodically
the pool is swapped for an out-of-distribution development case. The final
240-question benchmark in Table~\ref{tab:harness} is held out and never exposed
to the supervisor loop.

\textbf{A minimal harness.}
The initial agent is a ReAct-style model-in-a-loop \cite{yao2023react}
with a short list of terminal tools---%
\texttt{ls}, \texttt{read}, \texttt{grep}, a sandboxed non-persistent Python---and nothing else. CLEO builds on this minimal, file-system-native toolset, adding only the persistent per-case memory an analyst workflow needs.

\subsection{The optimization loop}
Inspired by \emph{autoresearch} \cite{karpathy2026autoresearch}, a
\emph{supervisor} model (a frontier LLM - we chose Opus-4.8~\cite{anthropicclaude}) drives an automatic
propose--test--keep loop over harness versions
(Algorithm~\ref{alg:loop}). A proposed change is kept only if it improves the current evaluation objective $a$---the pass-rate/soft-score summary produced by \textsc{Eval}---by a margin $\epsilon$ on the current graded subset $D_t$ \emph{and} survives our key contribution: a \textbf{justification check} that asks whether the edit is generic, auditable, and as lean as possible. This second gate is meant to guard against both
benchmark overfitting and scaffold bloat. Accepted changes define the
final harness: a small set of tools plus tuned guardrails,
retry/refusal handling, context compaction, memory policy, and tool-call
budgets. They trigger an error analysis that drives the
next proposal. To prevent the loop from fitting a single case, the evaluation set is periodically swapped for an out-of-distribution pool. \textsc{Eval} runs the candidate on 32 fixed and 12 randomly drawn Q\&A and scores each cited answer with an LLM judge \cite{zheng2023judge} against expert ground truth.
We optimize bottom-up rather than pruning a large off-the-shelf harness: in a
broad harness much must be removed, and deciding which capabilities are
unnecessary is itself a slow security and engineering exercise. In our
experiments, top-down optimization matched bottom-up on accuracy but barely
shrank the harness, whereas the near-empty CLEO loop added only what it
needed---e.g., read/write memory tools and a non-answer safety net with a
retry loop. Conversely, a \texttt{cite()} helper, proposed after error
analysis showed the judge penalizing answers with missing evidence, was
reverted: the agent rarely used it, so the change failed the justification
gate.

\begin{algorithm}[t]
\scriptsize
\caption{Autoresearch-style harness optimization.}
\label{alg:loop}
\begin{algorithmic}[1]
\Require model $M$, initial harness $h_0$ (CLEO v1), graded sets $D_t$, adoption margin $\epsilon > 0$ (min.\ accuracy gain to keep a change), supervisor $S$
\State $h \gets h_0$;\; $a \gets \textsc{Eval}(M,h,D_t)$
\Repeat
  \State $\delta \gets S.\textsc{Propose}(h)$ \Comment{add a tool / guardrail / loop tweak}
  \State $a' \gets \textsc{Eval}(M,\textsc{Apply}(\delta,h),\, D_t)$
  \If{$a' > a + \epsilon$ \textbf{and} $S.\textsc{Justified}(\delta,\mathrm{traces})$}
     \State $h \gets \textsc{Apply}(\delta,h)$;\; $a \gets a'$ \Comment{adopt}
  \Else
     \State discard $\delta$ \Comment{revert}
  \EndIf
  \State analyze; occasionally swap $D_t$ for an OOD pool
\Until{no justified gain}
\State \Return $h$
\end{algorithmic}
\end{algorithm}

\begin{table}[t]
\caption{Combined held-out benchmark results for four harnesses under
the same model}
\label{tab:harness}
\footnotesize
\setlength{\tabcolsep}{0.8pt}
\renewcommand{\arraystretch}{1.12}
\begin{tabular*}{\columnwidth}{@{\extracolsep{\fill}}p{0.43\columnwidth}cccc@{}}
\toprule
\textbf{Harness} & \textbf{LOC} & \textbf{Pass} & \textbf{Soft $\mu\pm\sigma$} & \textbf{Lat.} \\
\midrule
Claude Code SDK (bare) & $\sim$512k\tablefootnote{The SDK is closed-source and
distributed minified, so no exact source-line count is publicly verifiable. The
$\sim$512k figure is the one widely reported following the 2026 source-map
disclosure; we could not independently verify it.} & \textbf{85.0\%} & \textbf{8.48 $\pm$ 2.25} & 76s \\
PI \#1 (bare) & $\sim$75k & 80.8\% & 8.12 $\pm$ 2.70 & 126s \\
PI \#2 (CLEO's full prompts) & $\sim$75k & 83.8\% & 8.45 $\pm$ 2.39 & 121s \\
CLEO & \textbf{$\sim$3k} & 83.8\% & 8.41 $\pm$ 2.51 & 114s \\
\bottomrule
\end{tabular*}
\vspace{2pt}
\end{table}

\section{Results}
\paragraph{Evaluation setup.}
We evaluate four agent harnesses on our intelligence analysis benchmarks,
holding fixed the model (GLM-5.1~\cite{glm5}), and a single Claude Opus 4.8 LLM judge~\cite{anthropicclaude,zheng2023judge}. The judge receives the candidate answer and expert ground truth, then returns a correct/incorrect verdict and a 0--10 soft score.

We evaluate three baselines spanning two design axes: harness size (a large
off-the-shelf harness vs.\ a lightweight one) and prompt (generic vs.\
CLEO-optimized), reporting accuracy, latency, and estimated harness size for
each. The large harness is the Claude Code SDK~\cite{claudecodesdk}, a mature
production-grade system we treat as a reference point for what a
general-purpose harness achieves out of the box, paired with a generic
task-context prompt generated by Claude itself. The lightweight harness is
PI~\cite{pi}, a minimal execution loop with no domain-specific prompting---the
natural comparison for CLEO, though still considerably larger. PI~\#1 receives
the same prompt as Claude; PI~\#2 receives CLEO's full optimized prompt.

Table~\ref{tab:harness} shows that CLEO, despite its tiny harness, is
competitive with the substantially larger Claude Code SDK: an 83.8\% pass rate
and 8.41/10 soft score, within 1.2 percentage points and 0.07 soft-score points
of the best baseline. On 240 held-out questions that gap is roughly three
questions, so we read this as practical parity rather than a ranking. CLEO also
matches or outperforms both PI versions on pass rate with a much smaller harness
surface and similar latency. Given the small accuracy gap against the large LOC gap, CLEO is the best trade-off. The bottom-up
direction matters: we ran a top-down loop from PI, permitting all modifications and encouraging
removals, yet it matched bottom-up on accuracy while barely shrinking the harness.

\section{Conclusions}
This work shows that ground-up harness optimization is a practical direction for deploying agentic
systems in regulated domains: a minimal harness approaches the accuracy of the
strongest out-of-the-box harness with two orders of magnitude less code and
capability surface. The result holds within a defined scope: a base model
already capable of multi-step tool use, one investigative domain, and accuracy
scored by an LLM judge that domain experts and product managers spot-checked but
that we did not validate systematically. Within those limits, careful harness
design is an effective systems optimization: exposing only task-essential
capabilities retains competitive performance while improving maintainability,
auditability, and operational simplicity for production deployments.

\section*{Speaker Bio}
\textbf{Shai Nahum Gefen} is Data Science Team Lead at Dream Security, an AI-native company
for national cybersecurity and sovereignty, where she leads research on
agentic solutions and post training, and builds
LLM-based systems for national cyber defense.

\section*{Generative AI Disclosure}
Generative AI was used to assist with copy-editing and formatting of this
manuscript; all technical content, experiments, and analyses are the authors' own.

\balance


\begin{thebibliography}{99}

\bibitem{yao2023react}
S.~Yao, J.~Zhao, D.~Yu, N.~Du, I.~Shafran, K.~Narasimhan, and Y.~Cao.
\newblock ReAct: Synergizing Reasoning and Acting in Language Models.
\newblock In \emph{ICLR}, 2023.

\bibitem{yang2024sweagent}
J.~Yang, C.~E. Jimenez, A.~Wettig, K.~Lieret, S.~Yao, K.~Narasimhan, and O.~Press.
\newblock SWE-agent: Agent-Computer Interfaces Enable Automated Software Engineering.
\newblock In \emph{NeurIPS}, 2024.

\bibitem{hu2025adas}
S.~Hu, C.~Lu, and J.~Clune.
\newblock Automated Design of Agentic Systems.
\newblock In \emph{ICLR}, 2025.

\bibitem{karpathy2026autoresearch}
A.~Karpathy.
\newblock autoresearch: AI agents running experiments on single-GPU model training, automatically.
\newblock GitHub repository, 2026. \url{https://github.com/karpathy/autoresearch}.

\bibitem{glm5}
GLM-5 Team, A.~Zeng, X.~Lv, Z.~Hou, Z.~Du, Q.~Zheng, B.~Chen, D.~Yin, et al.
\newblock GLM-5: From Vibe Coding to Agentic Engineering.
\newblock arXiv:2602.15763, 2026.

\bibitem{zheng2023judge}
L.~Zheng, W.-L. Chiang, Y.~Sheng, et al.
\newblock Judging LLM-as-a-Judge with MT-Bench and Chatbot Arena.
\newblock In \emph{NeurIPS Datasets and Benchmarks}, 2023.

\bibitem{pi}
M.~Zechner.
\newblock Pi: A Minimal Coding Agent (four tools, sub-1k-token prompt).
\newblock GitHub, 2026. \url{https://github.com/badlogic/pi-mono}.

\bibitem{lewis2020rag}
P.~Lewis, E.~Perez, A.~Piktus, et al.
\newblock Retrieval-Augmented Generation for Knowledge-Intensive NLP Tasks.
\newblock In \emph{NeurIPS}, 2020.

\bibitem{li2026dci}
Z.~Li, H.~Zhang, C.~Wei, et al.
\newblock Beyond Semantic Similarity: Rethinking Retrieval for Agentic Search via Direct Corpus Interaction.
\newblock arXiv:2605.05242, 2026.

\bibitem{lin2026ahe}
J.~Lin, S.~Liu, C.~Pan, L.~Lin, S.~Dou, X.~Huang, H.~Yan, Z.~Han, and T.~Gui.
\newblock Agentic Harness Engineering: Observability-Driven Automatic Evolution of Coding-Agent Harnesses.
\newblock arXiv:2604.25850, 2026.

\bibitem{claudecodesdk}
Anthropic.
\newblock Claude Agent SDK (Claude Code).
\newblock GitHub, 2026. \url{https://github.com/anthropics/claude-agent-sdk-typescript}.

\bibitem{anthropicclaude}
Anthropic.
\newblock What's New in Claude Opus 4.8.
\newblock Claude Platform documentation, 2026. \url{https://platform.claude.com/docs/en/about-claude/models/whats-new-claude-4-8}.

\end{thebibliography}
\end{document}